\documentclass[letterpaper,twocolumn,10pt]{article}
\usepackage{usenix2019_v3}

\usepackage{tikz}
\usepackage{amsmath}

\usepackage{amsmath} 
\usepackage{amssymb}
\usepackage{subfigure}

\usepackage{algorithm}
\usepackage{algpseudocode}
\usepackage{amsmath}
\usepackage{graphics}
\usepackage{epsfig}

\usepackage{multirow}
\usepackage{booktabs}

\usepackage{appendix}
\usepackage{multicol}

\usepackage{graphicx}
\usepackage{array}

\begin{document}

\date{}

\title{\Large \bf Cloud-based Image Classification Service  Is Not Robust To Simple Transformations:A Forgotten Battlefield}

\author{
{\rm Dou Goodman }\\
Baidu X-Lab
\and
{\rm Tao Wei}\\
Baidu X-Lab
} 

\maketitle

\begin{abstract}
Many recent works demonstrated that Deep Learning models are vulnerable to adversarial examples.Fortunately, generating adversarial examples usually requires white-box access to the victim model, and the attacker can only access the APIs opened by cloud platforms. Thus, keeping models in the cloud can usually give a (false) sense of security.Unfortunately, cloud-based image classification service is not robust to simple transformations such as \emph{Gaussian Noise}, \emph{Salt-and-Pepper Noise}, \emph{Rotation} and \emph{Monochromatization}. In this paper,(1) we propose one novel attack method called \emph{Image Fusion(IF)} attack, which achieve a high bypass rate,can be implemented only with OpenCV and is difficult to defend; and (2) we make the first attempt to conduct an extensive empirical study of \emph{Simple Transformation} (ST) attacks against real-world cloud-based classification services. Through evaluations on four popular cloud platforms including Amazon, Google, Microsoft, Clarifai, we demonstrate that ST attack has a success rate of approximately 100\% except Amazon approximately 50\%, IF attack have a success rate over 98\% among different classification services. (3) We discuss the possible defenses to address these security challenges.Experiments show that our defense technology can effectively defend known ST attacks. 
\end{abstract}

\section{Introduction}
In recent years, Deep Learning(DL) techniques have been extensively deployed for computer vision tasks, particularly visual classification problems, where new algorithms reported to achieve or even surpass the human performance \cite{Liu2016SSD,Ren2015Faster,simonyan2014very,he2016deep,krizhevsky2012imagenet}. Success of DL algorithms has led to an explosion in demand. To further broaden and simplify the use of DL algorithms, cloud-based services offered by Amazon, Google, Microsoft, Clarifai, and others to offer various computer vision related services including image auto-classification, object identification and illegal image detection. Thus, users and companies can readily benefit from DL applications without having to train or host their own models.

\cite{szegedy2013intriguing} discovered an intriguing properties of DL models in the context of image classification for the first time. They showed that despite the state-of-the-art DL models are surprisingly susceptible to adversarial attacks in the form of small perturbations to images that remain (almost) imperceptible to human vision system. These perturbations are found by optimizing the input to maximize the prediction error and the images modified by these perturbations are called as adversarial example. The profound implications of these results triggered a wide interest of researchers in adversarial attacks and their defenses for deep learning in general.The initially involved computer vision task is image classification. For that, a variety of attacking methods have been proposed, such as L-BFGS of \cite{szegedy2013intriguing}, FGSM of \cite{goodfellow2014explaining}, PGD of \cite{madry2017towards},Deepfool of \cite{moosavi2016deepfool} ,C\&W of \cite{Carlini2016Towards} and so on.

Fortunately, generating adversarial examples usually requires white-box access to the victim model, and real-world cloud-based image classification services are more complex than white-box classifier, the architecture and parameters of DL models on cloud platforms cannot be obtained by the attacker. The attacker can only access the APIs opened by cloud platforms. Thus, keeping models in the cloud can usually give a (false) sense of security. Unfortunately, a lot of experiments have proved that attackers can successfully deceive cloud-based DL models without knowing the type, structure and parameters of the DL models\cite{YuanStealthy,Hosseini2017Google,Li2019Adversarial,goodman2019cloud,goodman2019transferability01,goodman2019hitbtransferability}.

In general, in terms of applications, research of adversarial example attacks against cloud vision services can be grouped into three main categories: query-based  attacks, transfer learning  attacks and spatial transformation  attacks. Query-based  attacks are typical black-box attacks, attackers do not have the prior knowledge and get inner information of  DL models through hundreds of thousands of queries to successfully generate an adversarial example \cite{Shokri2017Membership}.In \cite{ilyas2017query}, thousands of queries are required for low-resolution images. For high-resolution images, it still takes tens of thousands of times. For example, they achieves a 95.5\% success rate with a mean of 104342 queries to the black-box classifier. In a real attack, the cost of launching so many requests is very high.Transfer learning  attacks are first examined by \cite{szegedy2013intriguing}, which study the transferability between different models trained over the same dataset. \cite{Liu2016Delving} propose novel ensemble-based approaches to generate adversarial example . Their approaches enable a large portion of targeted adversarial example to transfer among multiple models for the first time.However, transfer learning attacks have strong limitations, depending on the collection of enough open source models, but for example, there are not enough open source models for pornographic and violent image recognition.

\begin{figure*}[t]
	\centering
	\subfigure[Origin image]{
		\label{fig:google:a}
		\begin{minipage}{0.3\linewidth}
			\centering
			\includegraphics[width=2in]{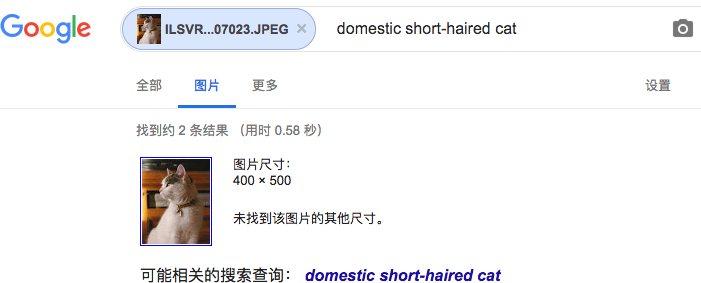}
		\end{minipage}
	}%
	\subfigure[Gaussian Noise]{
		\label{fig:google:b}
		\begin{minipage}{0.3\linewidth}
			\centering
			\includegraphics[width=2in]{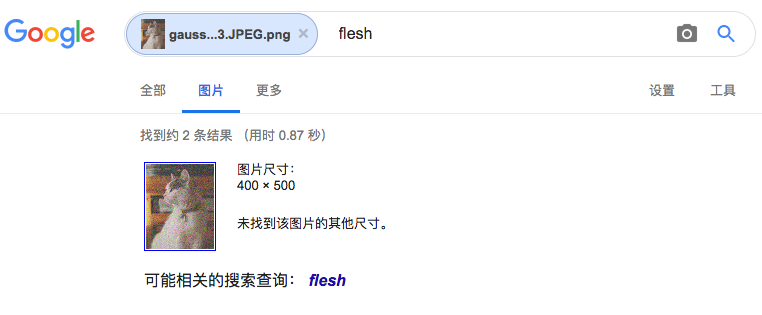}
		\end{minipage}
	}%
	\subfigure[Monochromatization]{
		\label{fig:google:c}
		\begin{minipage}{0.3\linewidth}
			\centering
			\includegraphics[width=2in]{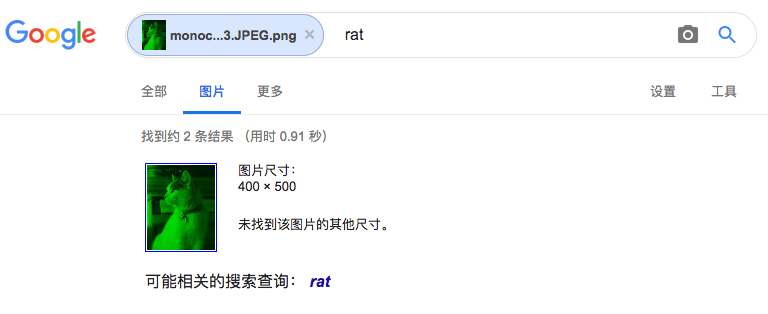}
		\end{minipage}
	}%
	
	\centering
	\caption{Illustration of the attack against Google Images Search.Figure \ref{fig:google:a} is origin image ,search result is a cat and Figure \ref{fig:google:b} is Gaussian Noise, search result is a flesh,Figure \ref{fig:google:c} is Monochromatization, search result is a rat. }
	\label{fig:google_image_search}
\end{figure*}

Spatial transformation is a different type of perturbation, as opposed to manipulating the pixel values directly as in prior works \cite{szegedy2013intriguing,moosavi2016deepfool,Carlini2016Towards,Hayes2017Machine,Brown2017Adversarial,ilyas2017query}. Perturbations generated through spatial transformation could result in large $L_p$ distance measures, but experiments show that such spatially transformed adversarial examples are perceptually realistic and more difficult to defend against with existing defense systems. This potentially provides a new direction in adversarial example generation and the design of corresponding defenses\cite{Xiao2018Spatially}.\cite{Hosseini2017Google} found that adding an average of 14.25\% impulse noise is enough to deceive the Google’s Cloud Vision API.\cite{YuanStealthy} found spatial transformation attacks can evade explicit content detection while still preserving their sexual appeal, even though the distortions and noise introduced are clearly observable to humans.

We further extend the spatial transformation attack, and choose the four methods which have the lowest cost to implement. These methods can be implemented without any Deep Learning knowledge and only need a few lines of OpenCV code, but the attack effect is very remarkable. Compared with previous work\cite{Xiao2018Spatially}, these methods are more threatening,we call it \emph{Simple Transformation} (ST) attacks,including \emph{Gaussian Noise}, \emph{Salt-and-Pepper Noise}, \emph{Rotation} and \emph{Monochromatization}.
To the best of our knowledge, no extensive empirical study has yet been conducted to black-box attacks and defences against real-world cloud-based image classification services. We summarize our main contributions as follows:
 
\begin{itemize}
	\item We propose one novel attack methods, Image Fusion(IF) attack ,which achieve a high bypass rate.Within our known range, there is no effective means of protection. 
	\item We make the first attempt to conduct an extensive empirical study of \emph{Simple Transformation}(ST) attacks against real-world cloud-based image classifier services. Through evaluations on four popular cloud platforms including Amazon, Google, Microsoft, Clarifai, we demonstrate that our ST attack has a success rate of approximately 100\%, IF attack have a success rate over 98\% among different classifier services.
	\item We discuss the possible defenses to address these security challenges in cloud-based classifier services.Our protection technology is mainly divided into model training stage and image preprocessing stage. Experiments show that our defense technology can effectively resist known ST attacks, such as  Gaussian Noise, Salt-and-Pepper Noise, Rotation, and Monochromatization.Through experiments, we prove how to choose different filters in the face of different noises, and how to choose the parameters of different filters.
\end{itemize}

\section{Theat model and criterion}
\subsection{Threat Model}
In this paper, we assume that the attacker can only access the APIs opened by cloud platforms, and get inner information of  DL models through limited queries to generate an adversarial example.Without any access to the training data, model, or any other prior knowledge, is a real black-box attack.
\subsection{Criterion and Evaluation}
The same with \cite{Li2019Adversarial} ,We choose top-1 misclassification as our criterion, which means that our attack is successful if the label with the highest probability generated by the neural networks differs from the correct label. 

We assume the original input is $O$, the adversarial example is $ADV$. For an RGB image $(m \times n \times 3)$, $(x, y, b)$ is a coordinate of an image for channel $ b (0 \leqslant  b \leqslant 2)$ at location $(x, y)$. 

We use Peak Signal to Noise Ratio (PSNR)\cite{Amer2002} to measure the quality of images.
\begin{equation}
PSNR = 10log_{10} (MAX^2/MSE)
\end{equation}
where $MAX =255$, $MSE$ is the mean square error.
\begin{equation}
MSE = \frac{1}{mn*3}*\sum_{b=0}^2\sum_{i=1}^n\sum_{j=1}^m ||ADV(i,j,b)-O(i,j,b)||^2
\end{equation}

Usually, values for the PSNR are considered between 20 and 40 dB, (higher is better) \cite{amer2005fast}.

We use structural similarity (SSIM) index to measure image similarity, the details of how to compute SSIM can be found in \cite{wang2004image}.Values for the SSIM are considered good between 0.5 and 1.0, (higher is better).

\section{Black-box attack algorithms}
\subsection{Problem Definition}
A real-world cloud-based image classifier service is a function $F(x) = y$ that accepts an input image $x$ and produces an output $y$. $F(.)$ assigns the label $C(x)=\arg \max_{i}F(x)_i$ to the input $x$. 

Original input is $O$, the adversarial example is $ADV$ and $\epsilon$ is the perturbation.

Adversarial example is defined as:

\begin{equation}
ADV=O+\epsilon
\end{equation}

We make a black-box  untargeted attack against real-world cloud-based classifier services $F(x)$:

\begin{equation}
C(ADV)\not=C(O)
\end{equation}

We also assume that we are given a suitable loss function $L( \theta ,x,y)$,for instance the cross-entropy loss for a neural network. As usual,  $\theta \in \mathbb{R}^p$ is the set of model parameters. 

\subsection{Simple Transformation}
Prior work such as \cite{Hosseini2017Google} only discussed Salt-and-Pepper Noise on Google vision APIs .\cite{YuanStealthy} report the first systematic study on the real-world adversarial images and their use in online illicit promotions  which belong to  image detectors. In the following, we explore the effect of 4 different ST  attacks on the classifier , including Gaussian Noise, Salt-and-Pepper Noise, Rotation, and Monochromatization.  All these image processing techniques are implemented with Python libraries, such as Skimage\footnote{https://scikit-image.org/} and OpenCV\footnote{https://opencv.org/}.

\begin{figure*}[thbp]
	\centering
	\subfigure[Origin]{
		\label{fig:cat:a}
		\begin{minipage}[t]{0.25\linewidth}
			\centering
			\includegraphics[width=1.5in]{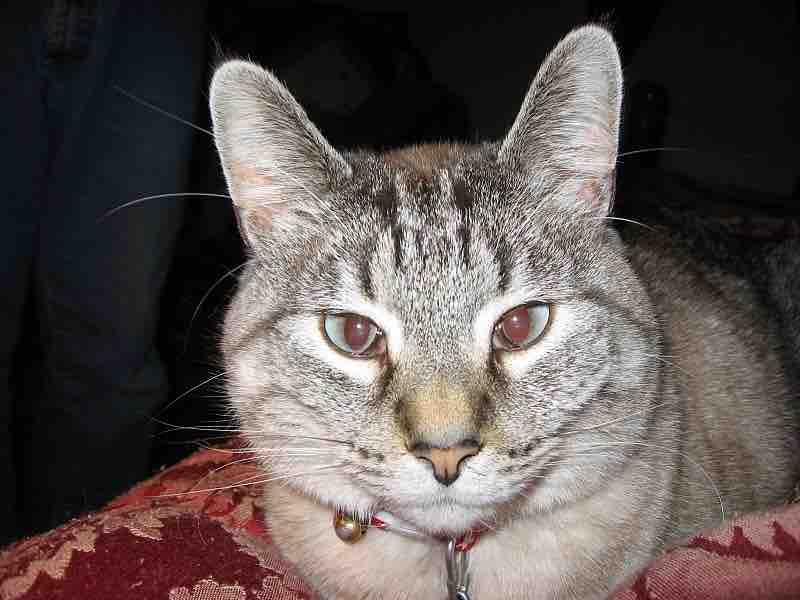}	
		\end{minipage}%
	}%
	\subfigure[$var=0.05$]{
		\label{fig:cat:b}
		\begin{minipage}[t]{0.25\linewidth}
			\centering
			\includegraphics[width=1.5in]{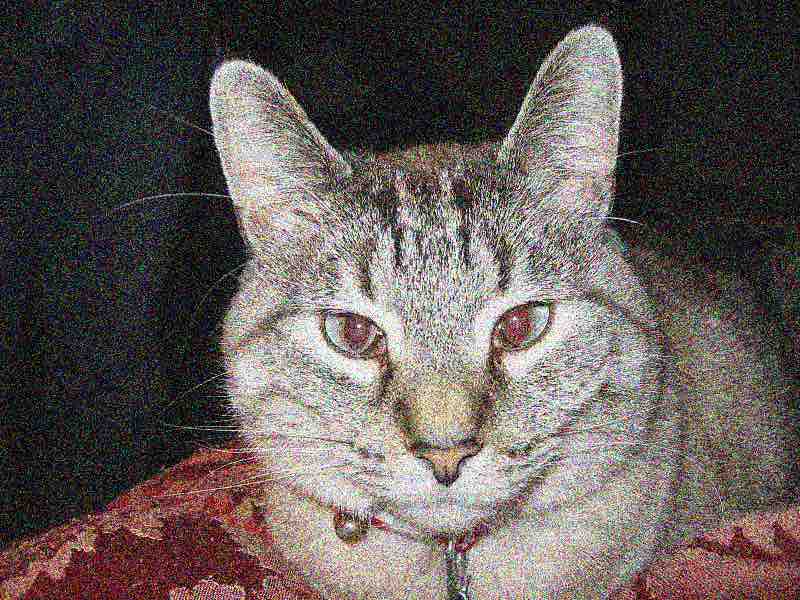}	
		\end{minipage}%
	}%
	\subfigure[$amount=0.01$]{
		\label{fig:cat:c}
		\begin{minipage}[t]{0.25\linewidth}
			\centering
			\includegraphics[width=1.5in]{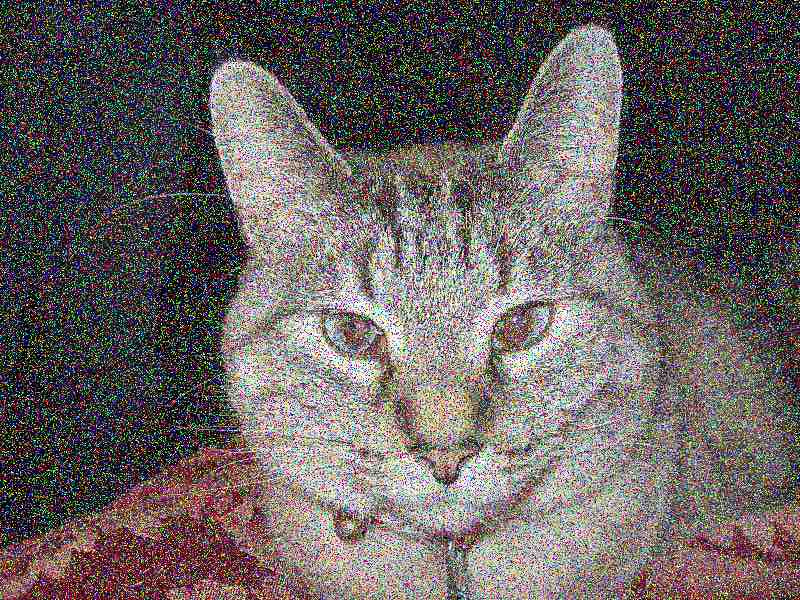}
			
		\end{minipage}
	}%
	\subfigure[$degree=45$]{
		\label{fig:cat:d}
		\begin{minipage}[t]{0.25\linewidth}
			\centering
			\includegraphics[width=1.5in]{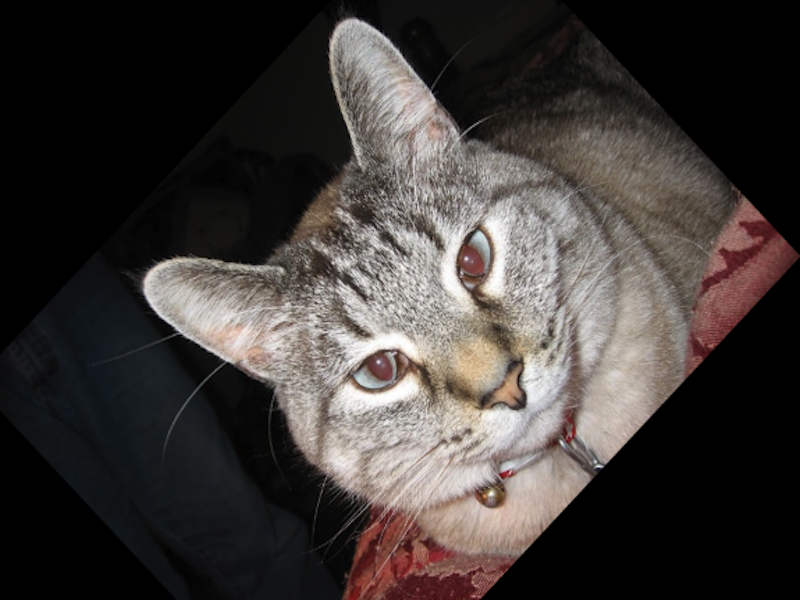}
			
		\end{minipage}
	}%
	
	\subfigure[$color=gray$]{
		\label{fig:cat:e}
		\begin{minipage}[t]{0.25\linewidth}
			\centering
			\includegraphics[width=1.5in]{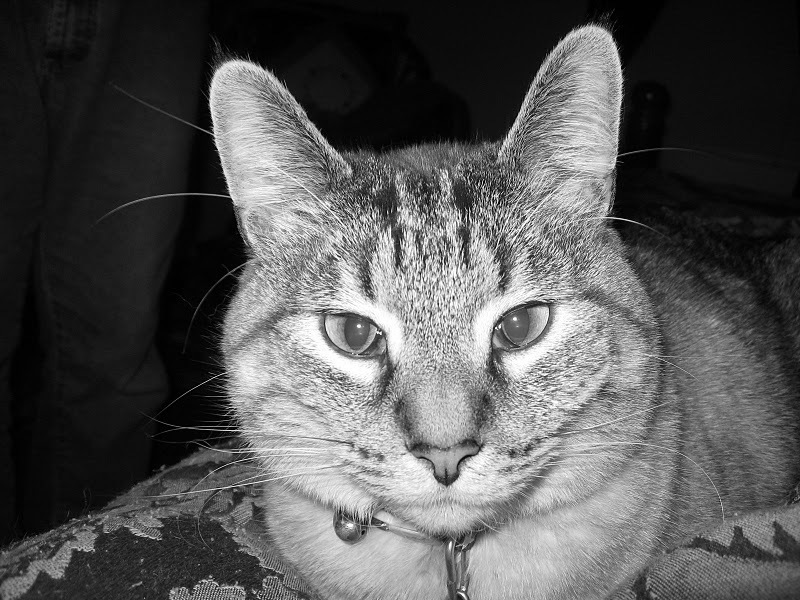}
			
		\end{minipage}
	}%
	\subfigure[$color=green$]{
		\label{fig:cat:f}
		\begin{minipage}[t]{0.25\linewidth}
			\centering
			\includegraphics[width=1.5in]{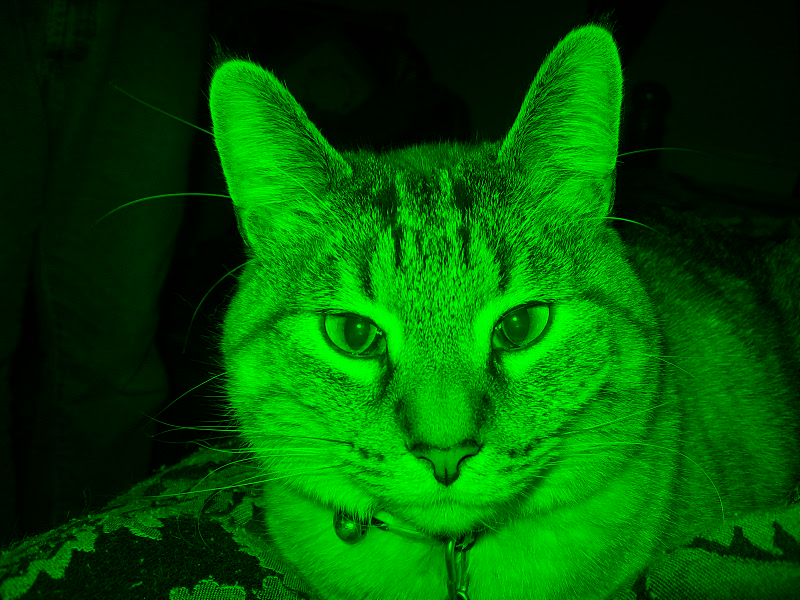}
			
		\end{minipage}
	}%
	\subfigure[$color=red$]{
		\label{fig:cat:g}
		\begin{minipage}[t]{0.25\linewidth}
			\centering
			\includegraphics[width=1.5in]{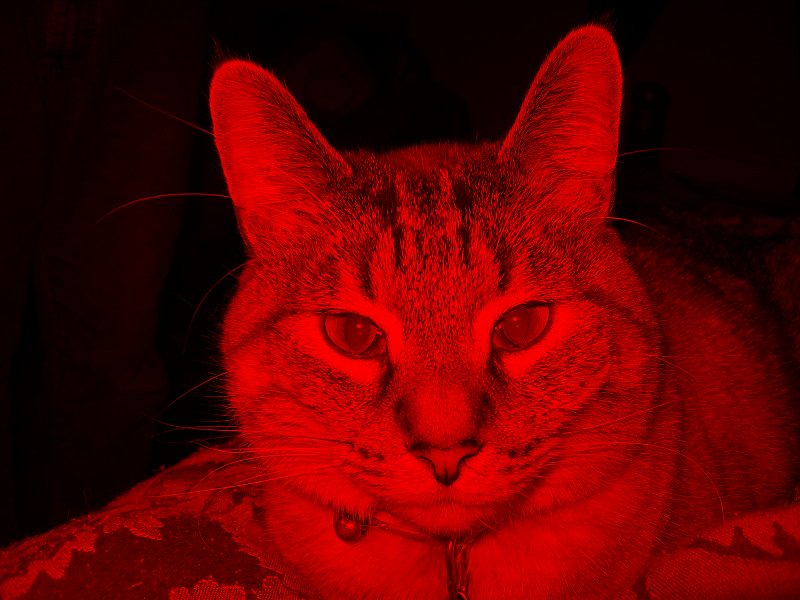}
			
		\end{minipage}
	}%
	\subfigure[$color=blue$]{
		\label{fig:cat:i}
		\begin{minipage}[t]{0.25\linewidth}
			\centering
			\includegraphics[width=1.5in]{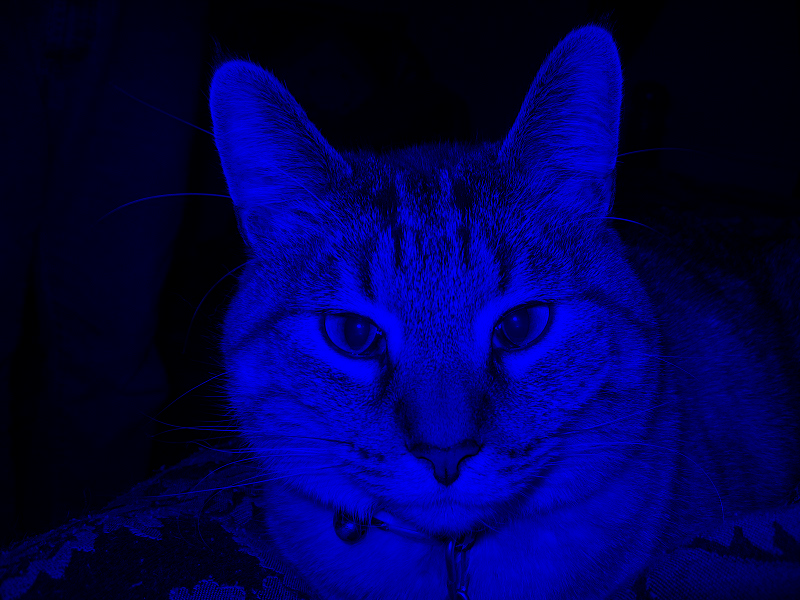}
			
		\end{minipage}
	}%
	\centering
	\caption{Illustration of ST attacks on a cat image.Figure \ref{fig:cat:a} is origin image,Figure \ref{fig:cat:b} is Gaussian Noise,Figure \ref{fig:cat:c} is Salt-and-Pepper Noise,Figure \ref{fig:cat:d} is Rotation and Figure \ref{fig:cat:e}\ref{fig:cat:f}\ref{fig:cat:g}\ref{fig:cat:i} is Monochromatization.}
\end{figure*}

\begin{figure*}[thbp]
	\centering
	\subfigure[Origin]{
		\label{fig:conv_cat:a}
		\begin{minipage}[t]{0.25\linewidth}
			\centering
			\includegraphics[width=1.5in]{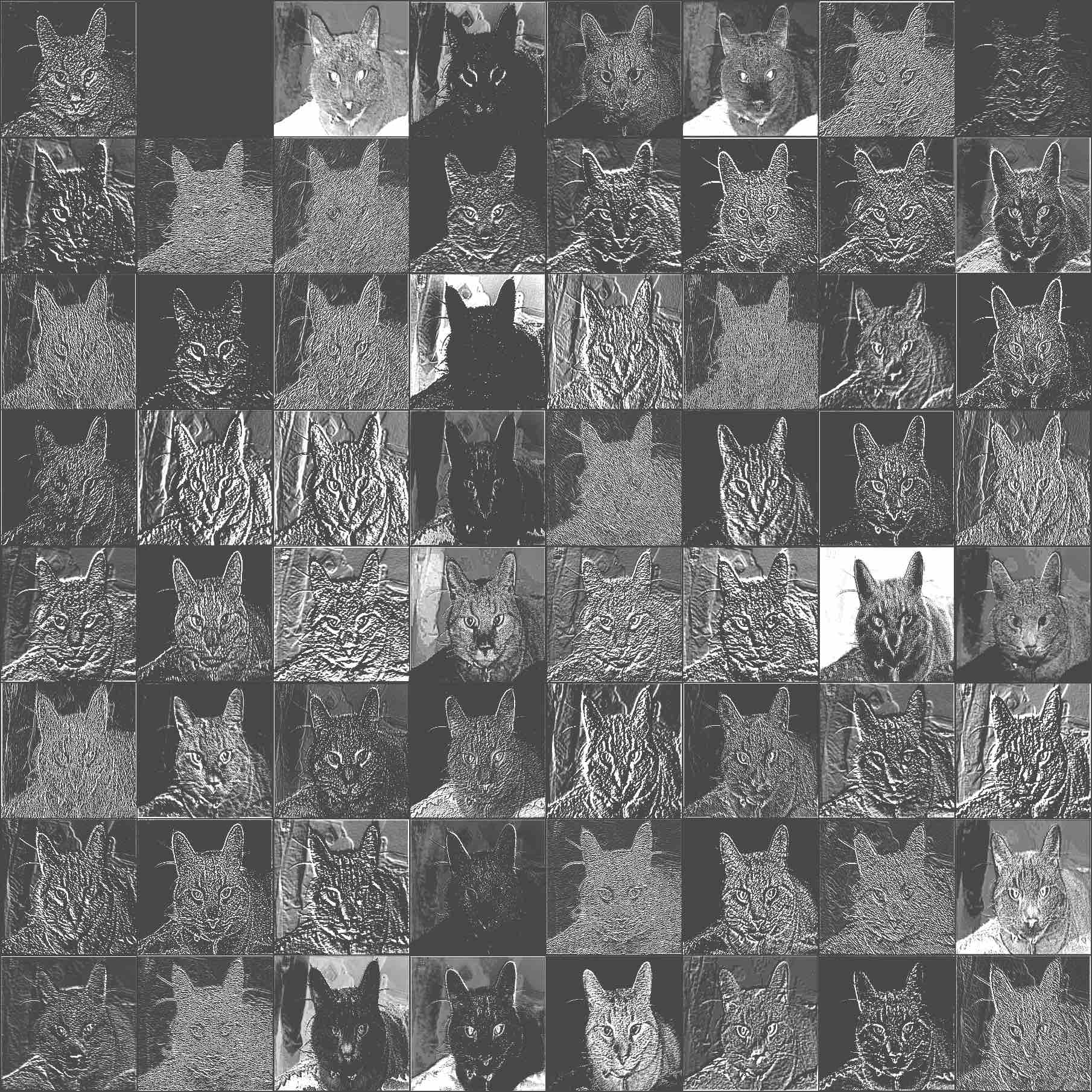}
			
		\end{minipage}%
	}%
	\subfigure[$var=0.05$]{
		\label{fig:conv_cat:b}
		\begin{minipage}[t]{0.25\linewidth}
			\centering
			\includegraphics[width=1.5in]{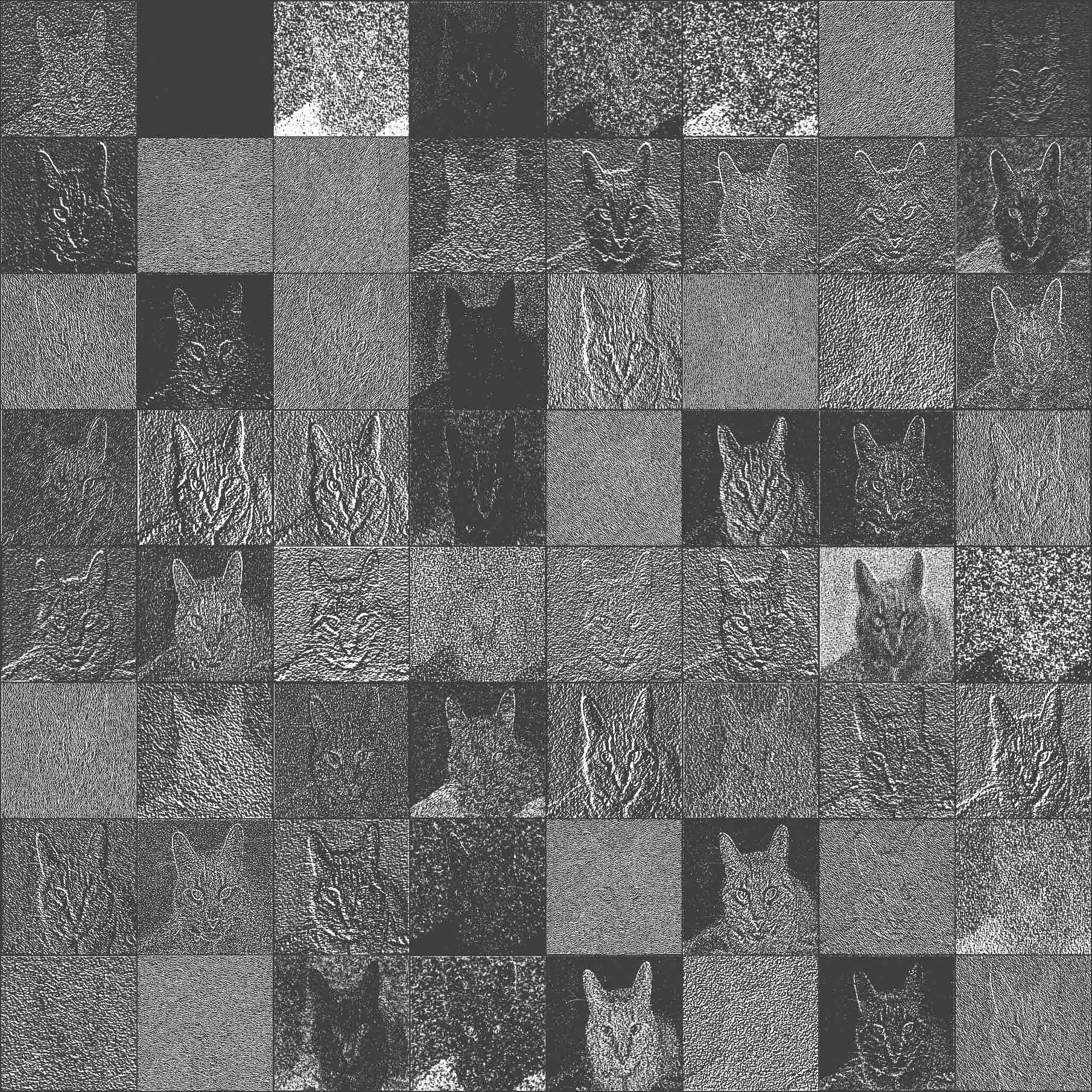}
			
		\end{minipage}%
	}%
	\subfigure[$amount=0.01$]{
		\label{fig:conv_cat:c}
		\begin{minipage}[t]{0.25\linewidth}
			\centering
			\includegraphics[width=1.5in]{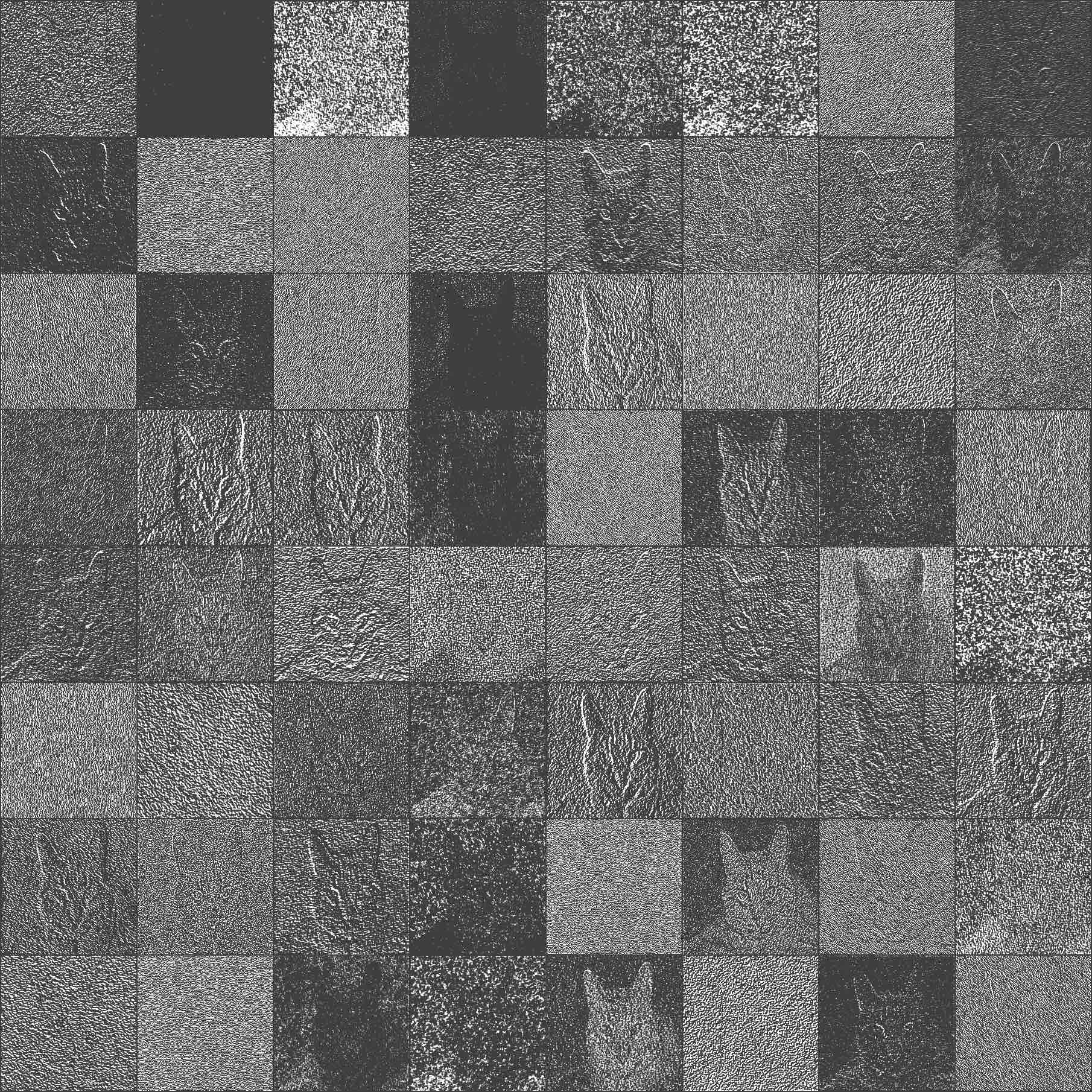}
			
		\end{minipage}
	}%
	\subfigure[$degree=45$]{
		\label{fig:conv_cat:d}
		\begin{minipage}[t]{0.25\linewidth}
			\centering
			\includegraphics[width=1.5in]{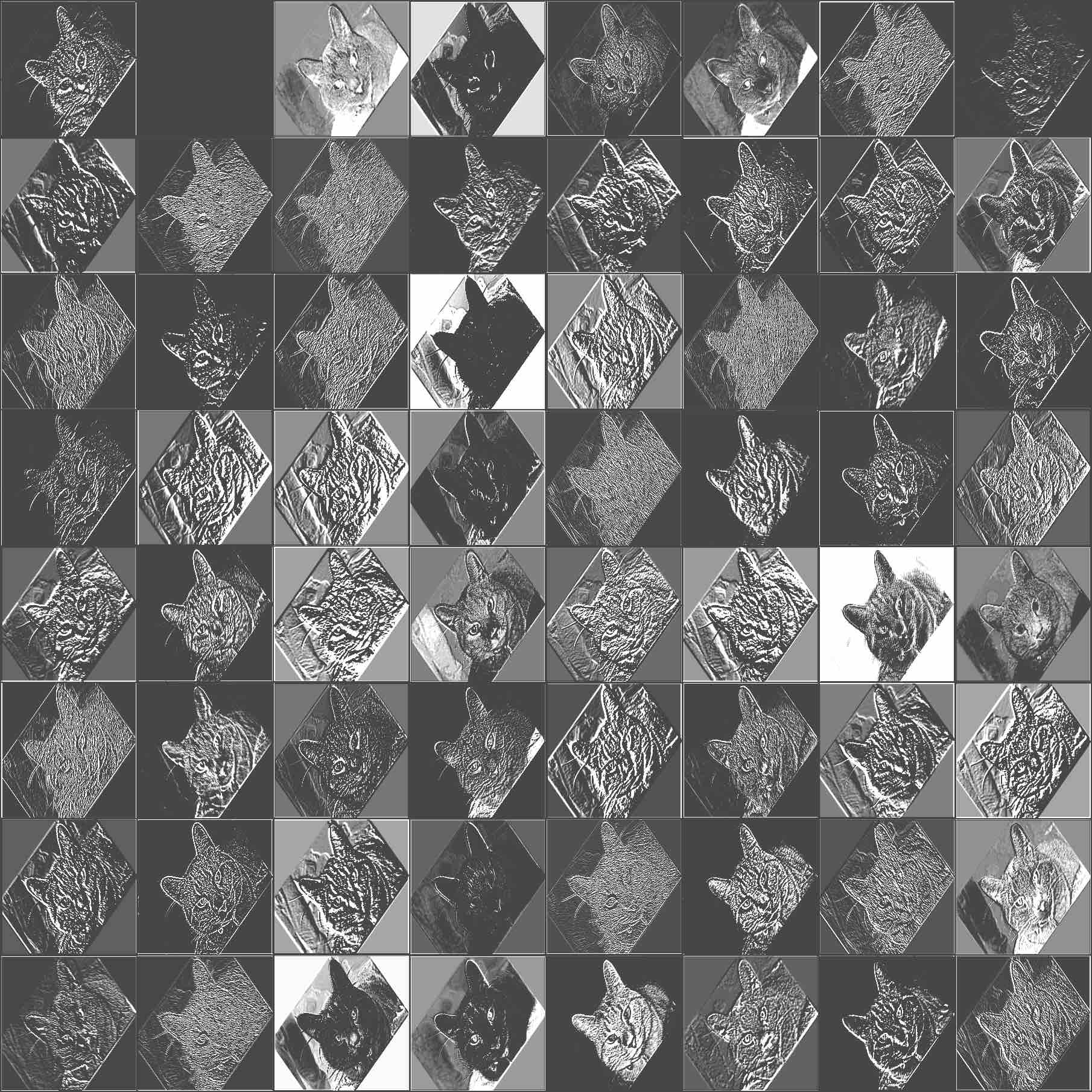}
			
		\end{minipage}
	}%

	\subfigure[$color=gray$]{
		\label{fig:conv_cat:e}
		\begin{minipage}[t]{0.25\linewidth}
			\centering
			\includegraphics[width=1.5in]{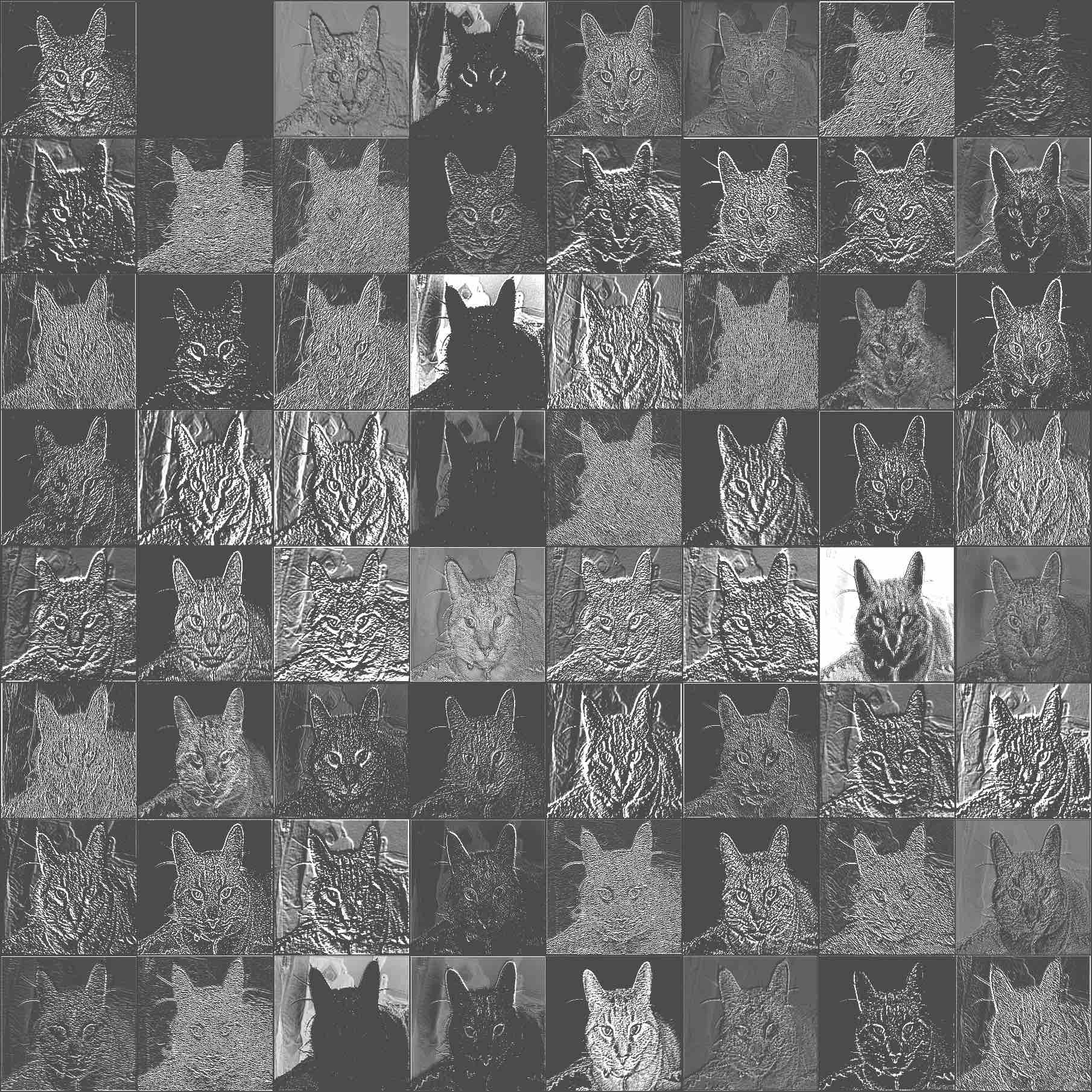}
			
		\end{minipage}
	}%
	\subfigure[$color=green$]{
		\label{fig:conv_cat:f}
		\begin{minipage}[t]{0.25\linewidth}
			\centering
			\includegraphics[width=1.5in]{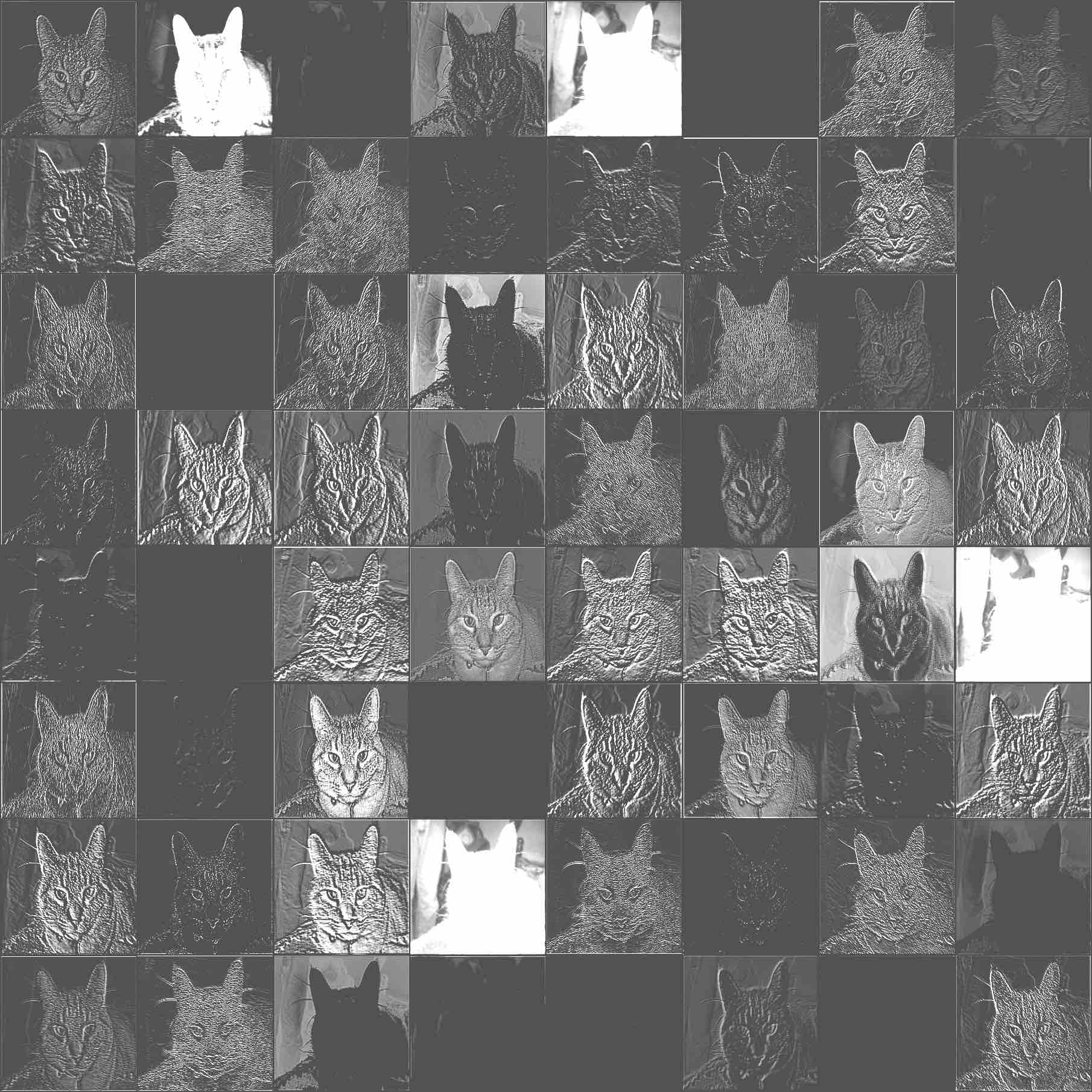}
			
		\end{minipage}
	}%
	\subfigure[$color=red$]{
		\label{fig:conv_cat:g}
		\begin{minipage}[t]{0.25\linewidth}
			\centering
			\includegraphics[width=1.5in]{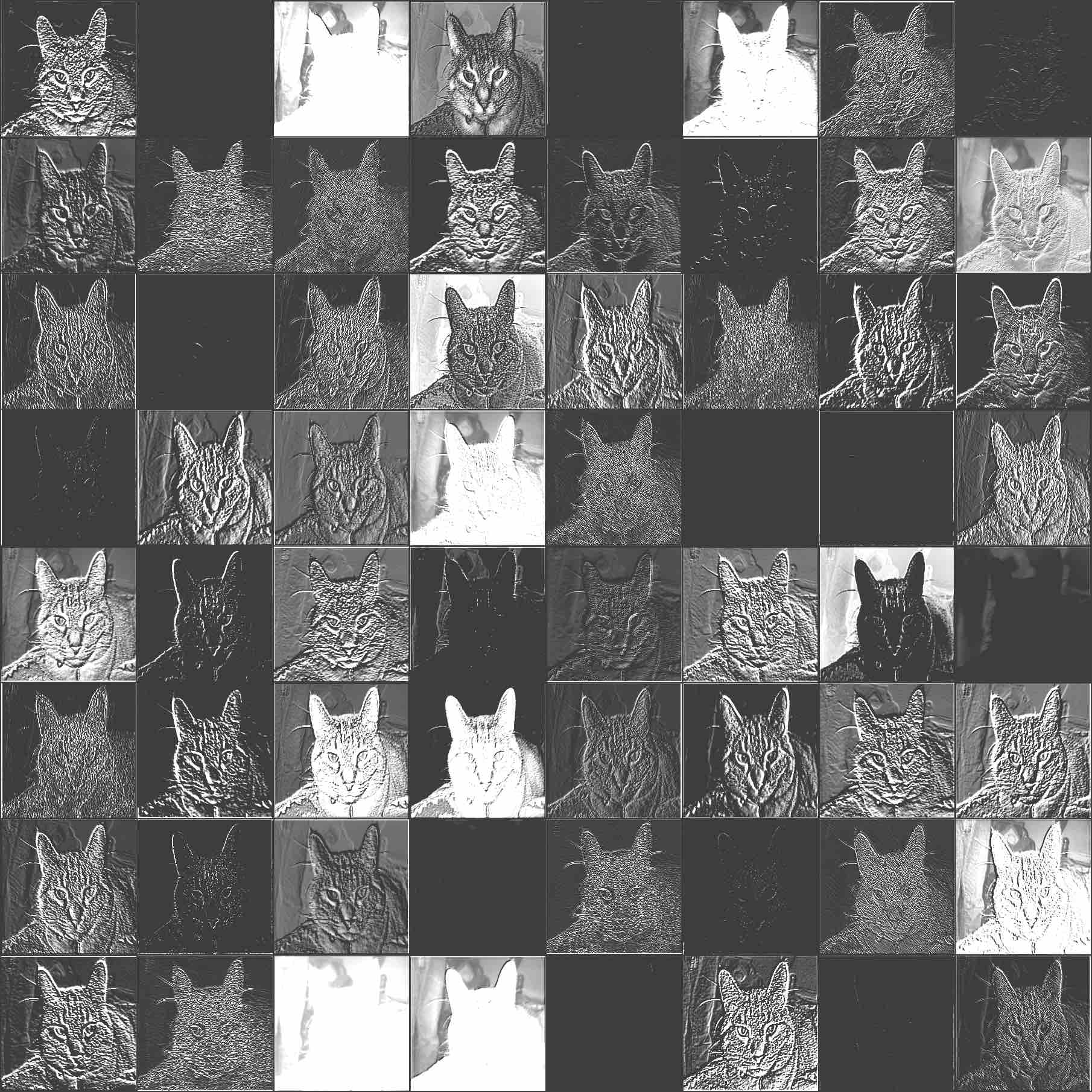}
			
		\end{minipage}
	}%
	\subfigure[$color=blue$]{
		\label{fig:conv_cat:i}
		\begin{minipage}[t]{0.25\linewidth}
			\centering
			\includegraphics[width=1.5in]{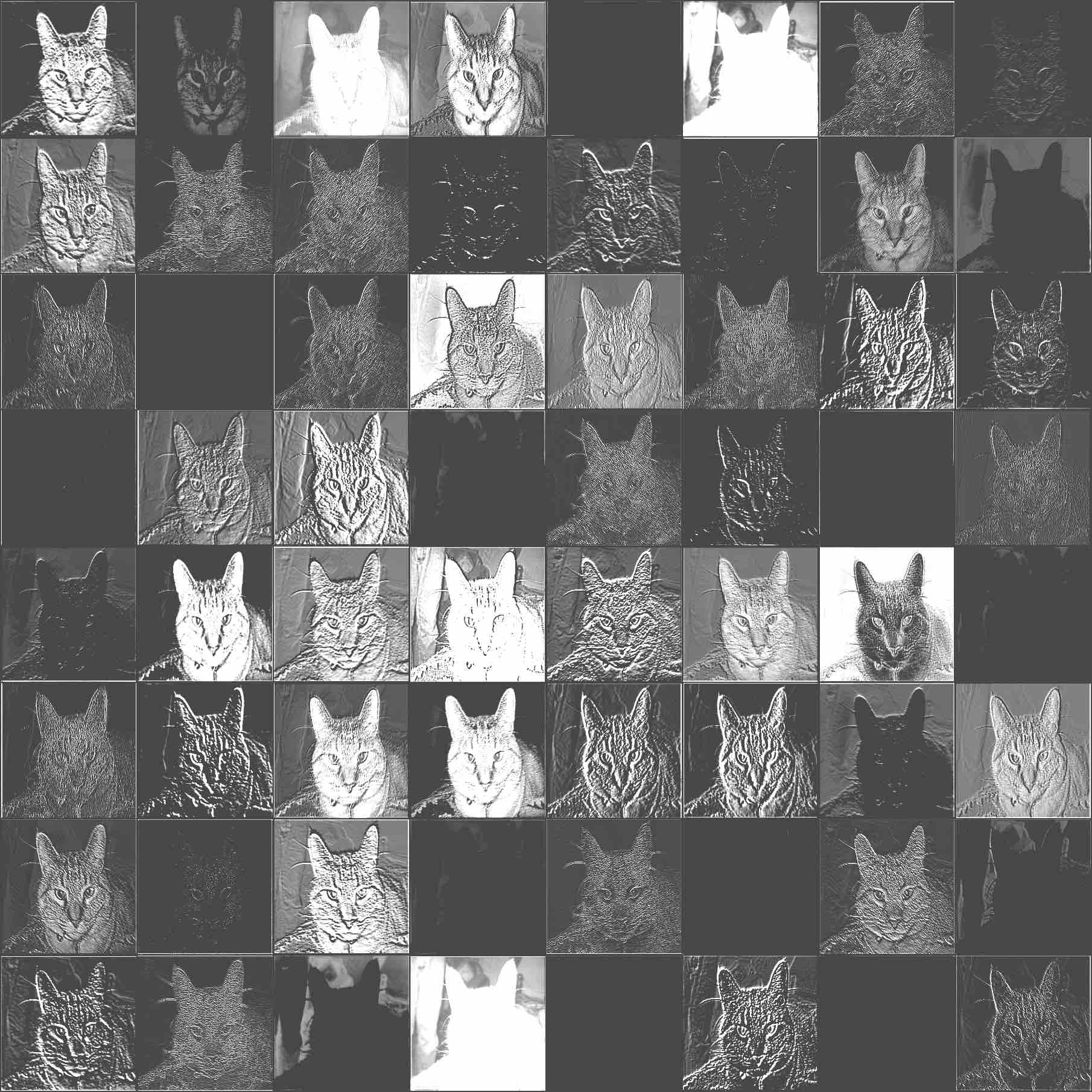}
			
		\end{minipage}
	}%
	\centering
	\caption{Features of distorted images extracted by VGG19’s first convolutional layer (after max pooling).Figure \ref{fig:conv_cat:a} is origin image,Figure \ref{fig:conv_cat:b} is Gaussian Noise,Figure \ref{fig:conv_cat:c} is Salt-and-Pepper Noise,Figure \ref{fig:conv_cat:d} is Rotation and Figure \ref{fig:conv_cat:e}\ref{fig:conv_cat:f}\ref{fig:conv_cat:g}\ref{fig:conv_cat:i} is Monochromatization.}
\end{figure*}

\subsubsection{Gaussian Noise and Salt-and-Pepper Noise}
Gaussian Noise is a kind of noise whose distribution satisfies Gaussian distribution. $N o i s e$ satisfies Gaussian distribution which $mean$ and $var$ represent the mean and variance,  Gaussian Noise is computed as follows:
\begin{equation}
A D V=clip(O+N o i s e),N o i s e \sim N\left(mean, var^{2}\right)
\end{equation}
For an RGB image $(m \times n \times 3)$, $(x, y, b)$ is a coordinate of an image for channel $ b (0 \leqslant  b \leqslant 2)$ at location $(x, y)$. Salt-and-Pepper Noise  is computed as follows:
\begin{equation}
A D V_{i, j, k}=\left\{\begin{array}{ll}{0} & {\text { with probability } \frac{amount}{2}} \\ {O_{i, j, k}} & {\text { with probability } 1-amount} \\ {255} & {\text { with probability } \frac{amount}{2}}\end{array}\right.
\end{equation}
Parameter $amount$ is between 0 to 1.0, which controls how many pixels become noise.
\subsubsection{Rotation}
Rotation refers to revolving the image clockwise, centering on the center of the original image, to get a new image, and filling the edges of the image with black. Parameter $degree$ controls the angle of rotation.

\subsubsection{Monochromatization}
Monochromatization refers to the preservation of only one channel data in RGB.

\begin{equation}
 AD V_{i, j, k}=\begin{cases}
O_{i, j, k} & k \text{ is the selected channel  }  \\ 
0 
\end{cases}
\end{equation}
Grayscale image is a special Monochromatization attack.

\begin{equation}
A D V=R*0.299 +G*0.587 +B*0.114
\end{equation}

\subsection{Image Fusion}
Image Fusion attack is a technique that applies a certain background image $B$ to the original image $O$. $B$ often contains rich high-frequency signals.So similar to noising, this technique makes image structural information difficult to extract.
\begin{equation}
ADV=\alpha*O+(1-\alpha)*B
\end{equation}
$\alpha$ is a hyperparameter between 0.0 and 1.0.

\begin{figure*}[h] 
	\centering 
	\includegraphics[width=1\textwidth]{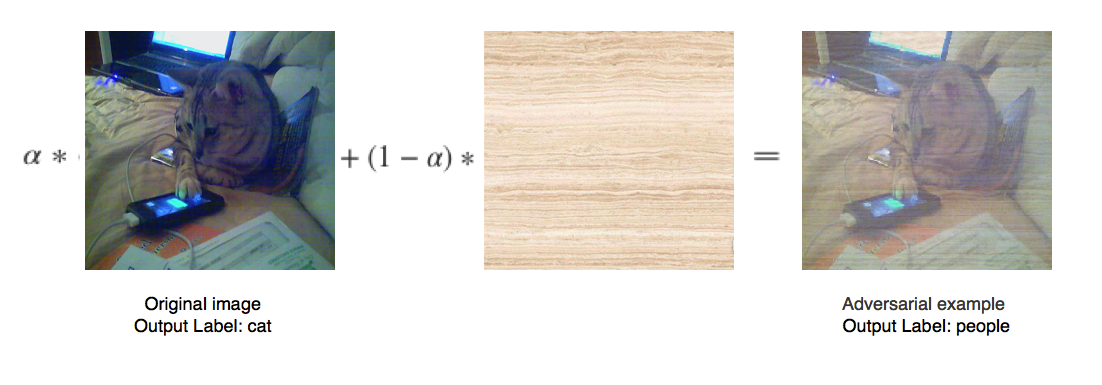}
	\caption{Illustration of the Image Fusion  attack on Clarifai Vision API,the hyperparameter of $\alpha$ is 0.5. By applying a certain background image $B$ to the original image $O$, we can force the API to output completely different labels. Captions are the labels with the highest confidence returned by the API.For adversarial example $ADV$, none of the output labels are related to corresponding original images. The cat image is chosen from the ImageNet val dataset. } 
	\label{fig:FL_demo} 
\end{figure*}

Figure \ref{fig:FL_demo} shows the API’s output label with the highest confidence score, for the original image $O$ and adversarial example $ADV$. As can be seen, unlike $O$, the API wrongly labels $ADV$, despite that the objects in $ADV$ are easily recognizable. 

We use the output of the first layer convolution of VGG19\cite{simonyan2014very} to visualize the process of image fusion attack in Figure \ref{fig:FL_conv_featuremap}. $\alpha$ is set to 0.2. We can find that although the proportion of the original image $O$ is only 0.2, the image can still be distinguished by the human naked eye as a cat. But in the  visualization of convolution feature map, we can find that the features extracted by convolution already contain a large amount information of background image $B$ , and the feature information of the original image has been seriously damaged.

\section{Defenses}

Defense adversarial examples is a huge system engineering, involving at least two stages: model training and image preprocessing.

\subsection{Model training}

\cite{goodfellow2014explaining} propose adversarial training to  improve the robustness of deep learning model. Retraining the model with new training data may be very helpful. In addition, we can find that ST attack is essentially a common image transformation method. It can improve the robustness of the model by adding \emph{Random Rotation}, \emph{Random Grayscale}, \emph{Random Horizontal Flip},\emph{Random Resize and Crop} and Noise filter in the data augmentation stage.We will focus on the introduction of \emph{Random Rotation} and \emph{Random Grayscale}. 

\subsubsection{Random rotation}

In the training process of the model, the robustness of the model against \emph{Rotation} attack can be improved by randomly rotating the image at a certain angle. The angle range of rotation is usually set as a parameter of random rotation.

\subsubsection{Random grayscale}

Adding gray images to training samples can improve the model's ability to resist \emph{Monochromatization} attack. Generally, the probability of randomly adding gray image is set to $p$.

\subsection{Image preprocessing}

The system’s robustness can be readily improved by applying a noise filter on the inputs to defense Gaussian Noise and Salt-and-Pepper Noise attacks, without the need for updating the image analysis algorithms. 

\subsubsection{Noise filter}
\emph{Gauss Filter} and \emph{Median Filter} are the most commonly used noise filters. We focus on these two filters.The industry often uses these two filters because they have stable and efficient implementations in OpenCV.

\emph{Gauss Filter} is a linear smoothing filter, which is suitable for removing gaussian noise and is widely used in image processing. Generally speaking, Gauss filtering is the process of weighted averaging of the whole image. The value of each pixel is obtained by weighted averaging of its own and other pixel values in its neighborhood. The specific operation of \emph{Gauss Filter} is to scan every pixel in the image with a template (or convolution, mask), and replace the value of the central pixel of the template with the weighted average gray value of the pixels in the neighborhood determined by the template.Generally, the size of convolution kernel $ksize$ is used as the parameter of \emph{Gauss Filter}.\emph{Gauss filter} is suitable for processing gaussian noise

\emph{Median Filter} is a non-linear smoothing technique. It sets the gray value of each pixel to the median value of the gray value of all pixels in a neighborhood window of the point, so that the surrounding pixels are close to the true value, thus eliminating isolated noise points.The size of window $ksize$ is used as the parameter of \emph{Median Filter}.\emph{Median Filter} is suitable for processing Salt-and-Pepper Noise.

\subsubsection{Grayscale}

Grayscale can  resist monochromatization attack, which refers to the preservation of only one channel data in RGB or the direct conversion of images into gray-scale images.Usually the picture is a color or gray image, gray image has only one color channel, and color image usually has three RGB color channels. Although the attack image has three color channels, only one channel has data, so it can be directly judged as cheating, or directly converted into a gray image of a channel.

\section{Experimental evaluation}

\subsection{Simple Transformation and Image Fusion Attacks}

\subsubsection{Datasets}

100 cat images  and 100 other animal images are selected from the ImageNet val set.Every input image is clipped to the size of $224 \times 224 \times 3$, where 3 is the number of RGB channels. The RGB value of the image is between 0 and 255.We use these 100 images of cats as original images to generate adversarial examples and make a black-box untargeted attack against real-world cloud-based image classifier services .We choose top-1 misclassification as our criterion, which means that our attack is successful if the label with the highest probability generated by the cloud-based image classifier service differs from the correct label "cat". We count the number of top-1 misclassification to calculate the escape rate.

\begin{table}[h]
	\caption{Correct label by cloud APIs}
	\label{tab:o}
	\centering
	\begin{tabular}{cccc}
		\toprule
		Platforms & Cat Images& Other Animals& All Images \\
		\midrule
		Amazon & 99/100 & 98/100 & 197/200\\
		Google & 97/100 & 100/100 &197/200 \\
		Microsoft & 58/100 & 98/100& 156/200 \\
		Clarifai & 97/100 & 98/100 &195/200 \\
		\bottomrule
	\end{tabular}
\end{table}

According to Table \ref{tab:o}, we can learn that Amazon and Google, which label 98.5\% of all  images correctly, have done a better job than other cloud platforms. 

\subsubsection{Simple Transformation }
To further understand the effectiveness of ST attacks, we apply 4 ST attacks such as Gaussian Noise,Salt-and-Pepper Noise, Rotation and Monochromatization,  each with 4 different settings (such as angle for rotation, density of noise, etc., see Table \ref{tab:st}). 
\begin{table}[htbp]
	\caption{Parameters used in 4 types of distortions}
	\label{tab:st}
	\centering
	
	\setlength{\tabcolsep}{1mm}{
		
		\begin{tabular}{ccccc}
			\toprule
			Method & L1& L2 & L3&L4\\
			\midrule
			Gaussian Noise(var)& 0.05&0.10  &0.15 &0.20\\
			Rotation(degree)&45 & 90 &135 &180\\
			Salt-and-Pepper Noise(amount)& 0.01 & 0.02&0.03 &0.04 \\
			Monochromatization(color)& blue & green & red&gray\\
			\bottomrule
		\end{tabular}
	}
\end{table}

\begin{figure*}[htbp]
	\centering
	
	\subfigure[Amazon]{%
			\includegraphics[width=3in]{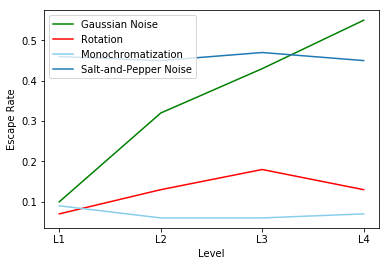}	
			\label{fig:st:a}
	}%
	\subfigure[Google]{%
			\includegraphics[width=3in]{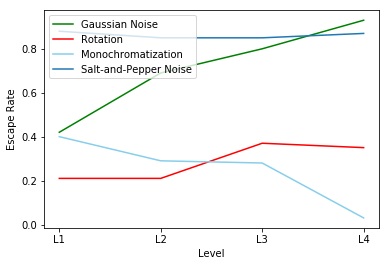}	
			\label{fig:st:b}
	}%

	\subfigure[Microsoft]{%
			\includegraphics[width=3in]{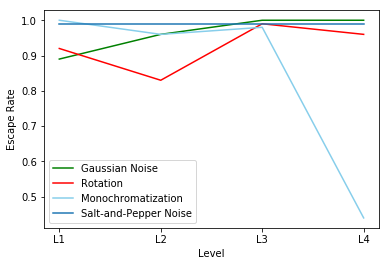}
			\label{fig:st:c}
	}%
	\subfigure[Clarifai]{%
			\includegraphics[width=3in]{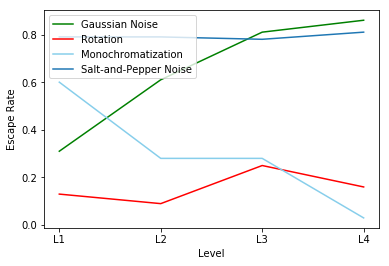}	
			\label{fig:st:d}
	}%

	\centering
	\caption{The escape rates of four ST attacks against Amazon,Google,Microsoft and Clarifai }
	\label{fig:st}
\end{figure*}

The escape rates of four ST attacks against Amazon,Google,Microsoft and Clarifai are shown in Figure \ref{fig:st}.

As we can see from Figure \ref{fig:st}, with proper settings, these four ST attacks are able to effectively degrade the performance of cloud-based image classification services including Amazon, Google, Microsoft, Clarifai. Gaussian Noise and Salt-and-Pepper Noise attacks have a success rate of approximately 100\% except Amazon. According to Figure \ref{fig:st}, we can learn that Amazon , which label above 50\% of all  images  of 4 ST attacks  correctly, have done a better job than other cloud platforms.  We speculate that Amazon has done a lot of work in image preprocessing to improve the robustness of the whole service.

\subsubsection{Image Fusion}

\begin{figure*}[htbp]
	\centering
	
	\subfigure[PSNR]{
		\label{fig:fl_conv:a}
		\includegraphics[width=3in]{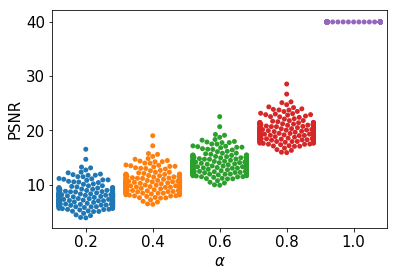}
	}%
	\subfigure[SSIM]{
		\label{fig:fl_conv:b}
		\includegraphics[width=3in]{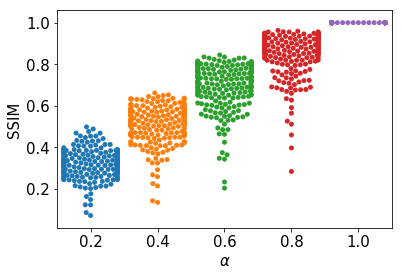}
	}%
	
	\centering
	\caption{We show the PSNR and SSIM of \emph{Image Fusion} attack.The images are from our val datasets.$\alpha$ controls image quality and similarity,image quality and similarity improve as the parameter increases.When $\alpha=1$, it is equivalent to $ADV=O$. The fusion image is exactly the same as the original image, so $SSIM = 1$ and PSNR is infinity. For display convenience, PSNR is uniformly expressed as 40.}
	\label{fig:IF_psnr_ssim}
\end{figure*}

The image fusion parameter $\alpha$ represents the proportion of the original image $O$ in the newly generated image $ADV$.When $\alpha=1$, it is equivalent to $ADV=O$. The fusion image is exactly the same as the original image, so $SSIM = 1$ and PSNR is infinity.We launch \emph{Image Fusion} attack against the images in the val dataset, and the fusion parameter $\alpha$ range from 0.2 to 1.0. As shown in the Figure \ref{fig:IF_psnr_ssim}, we can find that $\alpha$ controls image quality and similarity,image quality and similarity improve as the parameter increases.

\begin{figure}[h] 
	\centering 
	\includegraphics[width=3in]{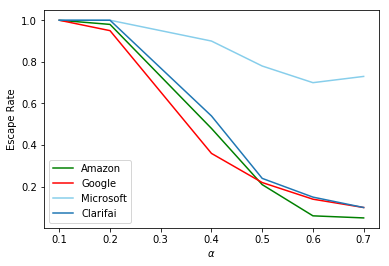}
	\caption{We increase $\alpha$  from 0.1 to 0.7, the figure records the escape rates of  IF attack against cloud-based image classification services under different $\alpha$ } 
	\label{fig:if} 
\end{figure}

The escape rates of \emph{Image Fusion} attack are shown in Figure \ref{fig:if}. From Figure \ref{fig:if}, we know that all the cloud-based image classifier services are vulnerable to \emph{Image Fusion} attacks . $\alpha$ controls the escape rate and  the escape rate decreases as the parameter increases , when $\alpha$ is between 0.1 and 0.2,\emph{Image Fusion} attack has a success rate of approximately 100\%.

\subsection{Defenses}

\subsubsection{Datasets}

In the experimental stage, it is unrealistic to obtain the image classification model of cloud service providers directly and retrain it. In order to simulate the real environment as much as possible, we used VGG19 model to retrain our image classification model on the train and val dataset of ImageNet.Detailed description of data set in Table \ref{tab:imagenet}.

\begin{table}[htbp]
	\caption{Training and val images of our defenses. A total of four kinds of animal pictures in the ImageNet data set are used. The training set size of each animal is 1300 and the val set size is 50.}
	\label{tab:imagenet}
	\centering
	\begin{tabular}{ccccc} 
		\toprule
		Label ID & Class Name & Train images&Val images \\
		\midrule
		0& tench&  1300&50 \\
		1& goldfish & 1300&50 \\
		2& white shark & 1300&50 \\
		283& cat & 1300&50 \\
		\bottomrule
	\end{tabular}
\end{table}

\subsubsection{Data augmentation and noise filter}

We choose top-1 accuracy as our criterion and high accuracy means effective defense. We 
define top-1 accuracy as our defense rates.Every input image is clipped to the size of $224 \times 224 \times 3$, where 3 is the number of RGB channels. The RGB value of the image is between 0 and 255.

During the training phase, we use data augmentation techniques and a noise filter on the inputs during the image preprocessing phase.\emph{Std} is short for \emph{Standard VGG19 retrained by us},\emph{DA} is short for \emph{Data Augmentation} and \emph{NF} is short for \emph{Noise Filter}.\emph{Std+DA+NF} means that we use data augmentation during the training phase and a noise filter on the inputs during the image preprocessing phase.Top-1 accuracy of \emph{Std} is $99.5\%$,while \emph{Std+DA} is $98.5\%$.It can be considered that the accuracy of the model after data enhancement is acceptable. 

We focus on the parameters of the noise filter in image preprocessing.We use the image in the val dataset to launch \emph{Gaussian Noise} attack and \emph{Salt-and-Pepper Noise}, and then denoise by filter. The parameter of \emph{Gaussian Noise} attack are $var$, parameter of \emph{Salt-and-Pepper Noise} attack is $amount$,and the parameter of the noise filter are $ksize$.

\begin{figure}[htbp]
	\centering
	
	\subfigure[PSNR]{
		\label{fig:nf_2:a}
		\includegraphics[width=3in]{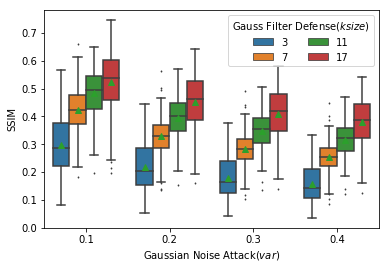}
	}%

	\subfigure[SSIM]{
		\label{fig:nf_2:b}
	    \includegraphics[width=3in]{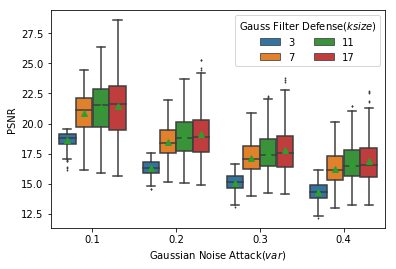}
	}%

	\centering
	\caption{We increase $var$ of  \emph{Gaussian Noise Attack} from 0.1 to 0.4,  $ksize$ of the \emph{Gauss Filter} from 3 to 17, the Figure \ref{fig:nf_2:a} records the PSNR  and  the Figure \ref{fig:nf_2:b} records the SSIM  under different parameters.When the parameters $var$ of \emph{Gaussian Noise Attack} are the same, increasing the window size parameters $ksize$ of the \emph{Gauss Filter} will improve the image similarity and the image quality.}
	\label{fig:nf_2}
\end{figure}

We increase $var$ of  \emph{Gaussian Noise Attack} from 0.1 to 0.4,$ksize$ of the \emph{Gauss Filter} from 3 to 17, the Figure \ref{fig:nf_2:a} records the PSNR  and  the Figure \ref{fig:nf_2:b} records the SSIM  under different parameters.When the parameters $var$ of \emph{Gaussian Noise Attack} are the same, increasing the window size parameters $ksize$ of the \emph{Gauss Filter} will improve the image similarity and the image quality.

\begin{figure*}[thbp]
	\centering
	\subfigure[]{
		\label{fig:noise_defense:a}
		\begin{minipage}{0.5\linewidth}
			\centering
			\includegraphics[width=3in]{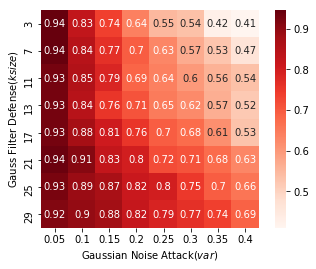}
		\end{minipage}
	}%
	\subfigure[]{
		\label{fig:noise_defense:b}
		\begin{minipage}{0.5\linewidth}
			\centering
			\includegraphics[width=3in]{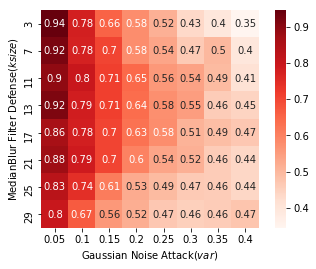}
		\end{minipage}
	}%
	
	\subfigure[]{
		\label{fig:noise_defense:c}
		\begin{minipage}{0.5\linewidth}
			\centering
			\includegraphics[width=3in]{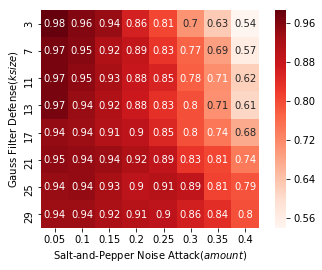}
		\end{minipage}
	}%
	\subfigure[]{
		\label{fig:noise_defense:d}
		\begin{minipage}{0.5\linewidth}
			\centering
			\includegraphics[width=3in]{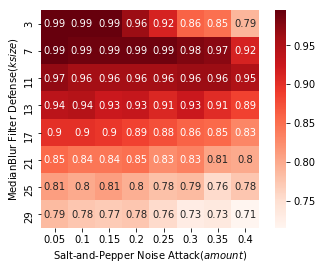}
		\end{minipage}
	}%
	
	\centering
	\caption{Defense rate under different noise and different noise filter parameters. }
	\label{fig:noise_defense}
\end{figure*}

From Figure \ref{fig:noise_defense:a}, we can see that facing the same \emph{Gaussian Noise}, the larger the window size $ksize$ of the \emph{Gauss Filter}, the better the filtering effect and the higher the defense rate.From Figure \ref{fig:noise_defense:a} and Figure \ref{fig:noise_defense:b}, we can see that when the attack parameters of \emph{Gaussian Noise} are the same, the defense rate of \emph{Gauss Filter} is higher than that of \emph{Median Filter}, and the \emph{Gauss Filter} with $ksize=29$  has the best protective effect against different values of \emph{Gaussian Noise}.

From Figure \ref{fig:noise_defense:c} and Figure \ref{fig:noise_defense:d}, we can see that when the attack parameters of \emph{Salt-and-Pepper Noise} are the same, the defense rate of \emph{Median Filter} is higher than that of \emph{Gauss Filter}, and the \emph{Median Filter} with $ksize=11$ has the best protective effect against different values of \emph{Salt-and-Pepper Noise}.

\begin{figure*}[htbp]
	\centering
	
	\subfigure[Gaussian Noise]{
		\label{fig:defense:a}
			\centering
			\includegraphics[width=3in]{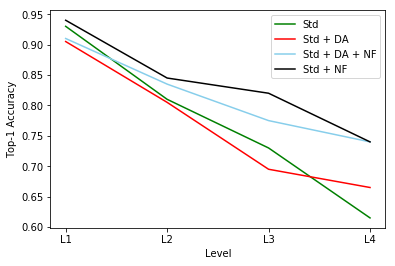}
	}%
	\subfigure[Rotation]{
		\label{fig:defense:b}
			\centering
			\includegraphics[width=3in]{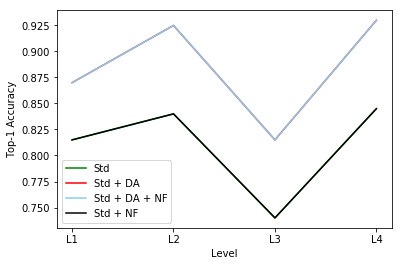}
	}%

	\subfigure[Salt-and-Pepper Noise]{
		\label{fig:defense:c}
			\centering
			\includegraphics[width=3in]{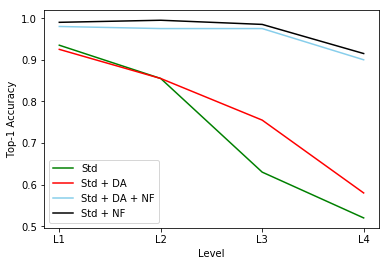}
	}%
	\subfigure[Monochromatization]{
		\label{fig:defense:d}
			\centering
			\includegraphics[width=3in]{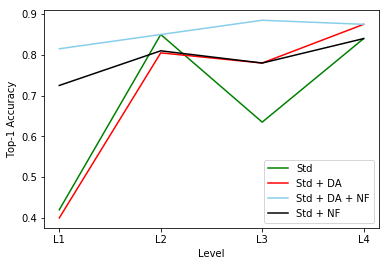}
	}%
	
	\centering
	\caption{The defense rates of four ST attacks.Figure \ref{fig:defense:a} show the defense rates of Gaussian Noise, Figure \ref{fig:defense:b} show the defense rates of Rotation,Figure \ref{fig:defense:c} show the defense rates of Salt-and-Pepper Noise and Figure \ref{fig:defense:d} show the defense rates of Monochromatization.Lines \emph{Std} and \emph{Std+NF},\emph{Std+DA} and \emph{Std+DA+NF} coincide in Figure \ref{fig:defense:b}.}
	\label{fig:defense}
\end{figure*}

\begin{table*}[htbp]
	\caption{Methods and parameters of defenses during the training and image preprocessing phase.}
	\label{tab:defense_parameters}
	\centering
	\begin{tabular}{ccc} 
		\toprule
		Stage & Method & Parameters \\
		\midrule
		\multirow{6}{*}{Training}& Random Rotation(degree range)&  (0,360)\\
						  & Random Grayscale(probability) & 0.5\\
						  & Random Horizontal Flip(probability) & 0.5\\
						  & Random Resize and Crop(image size) & 224\\
						  & Gauss Filter(ksize) & 29\\
						  & Median Filter(ksize) & 11\\
		\midrule
		\multirow{2}{*}{Image preprocessing}&Median Filter(ksize) & 11\\
						 & Grayscale & N/A\\
		\bottomrule
	\end{tabular}
\end{table*}

As can be seen from Figure \ref{fig:defense:a} and Figure \ref{fig:defense:c},the ability of the model to defense \emph{Gaussian Noise} and \emph{Salt-and-Pepper Noise}  attacks can be improved by using the noise filter.We can see from Figure \ref{fig:defense:b} that data augmentation improves the ability to defense \emph{Rotation} attacks.Simultaneous use of data enhancement and noise filter can defense \emph{Monochromatization} attacks,which can be seen from Figure \ref{fig:defense:d}.In addition, lines \emph{Std} and \emph{Std+NF},\emph{Std+DA} and \emph{Std+DA+NF} coincide in Figure \ref{fig:defense:b}.It can be seen that image denoising has no effect on rotation attack.

\begin{figure}[htbp] 
	\centering 
	\includegraphics[width=0.4\textwidth]{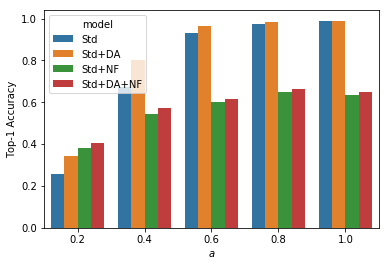}
	\caption{The defense rates of IF attacks under different $a$.There is almost no defensive effect improvement in the face of data augmentation and image preprocessing} 
	\label{fig:FL_defense} 
\end{figure}

Unfortunately, our IF attack, as shown in the Figure \ref{fig:FL_defense}, has almost no defensive effect improvement in the face of data augmentation and image preprocessing.

\section{Discussion}

\subsection{Effect of Attacks}

Our research shows that ST and IF attacks  can reduce the accuracy of mainstream image classification services in varying degrees. To make matters worse, for any image classification service, we can find a way that can be almost 100\% bypassed as shown in Table \ref{tab:discussion_attacks}. As shown in the Figure \ref{fig:google_image_search}, we can attack image search in the same way.

\begin{table}[htbp]
	\caption{Escape rates of cloud-based detectors attack.}
	\label{tab:discussion_attacks}
	\centering
	
	\setlength{\tabcolsep}{9mm}{
	
	\begin{tabular}{ccc}
		\toprule
		Platforms & ST& IF \\
		\midrule
		Amazon & 0.54 &0.98 \\
		Google & 0.98 & 0.98  \\
		Microsoft & 1.0& 1.0 \\
		Clarifai & 0.98 &0.98 \\
		\bottomrule
	\end{tabular}
	}
\end{table}

\subsection{Effect of Defenses}

Defense adversarial examples is a huge system engineering, involving at least two stages: model training and image preprocessing.Experiments show that our defense technology can effectively resist known ST attacks, such as Gaussian Noise, Salt-and-Pepper Noise, Rotation, and Monochromatization. 

\begin{table}[hbtp]
	\caption{Defense rates of ST attack.Our defense technology can raise the defense rate to more than 80\%,we have used the black line to thicken it.}
	\label{tab:discussion_defense}
	\centering
	\setlength{\tabcolsep}{1mm}{
	\begin{tabular}{ccc}
		\toprule
		Attack & Without Defense& With Defense \\
		\midrule
		Gaussian Noise & 0.60&\textbf{0.80}\\
		Rotation &0.70&\textbf{0.80}\\
		Salt-and-Pepper Noise &0.50&\textbf{0.95} \\
		Monochromatization & 0.4&\textbf{0.80} \\
		Image Fusion & 0.25&0.40 \\
		\bottomrule
	\end{tabular}
}
\end{table}

\begin{figure*}[h]
	\subfigure[Original image $O$]{
		\label{fig:fl_conv:a}
		\begin{minipage}[t]{0.3\linewidth}
			\centering
			\includegraphics[width=2in]{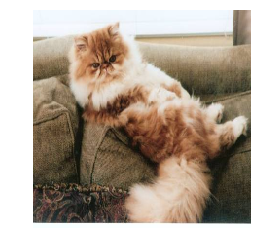}	
		\end{minipage}
	}%
	\subfigure[Backgroud $B$]{
		\label{fig:fl_conv:b}
		\begin{minipage}[t]{0.3\linewidth}
			\centering
			\includegraphics[width=2in]{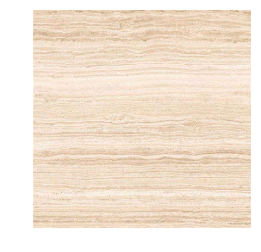}	
		\end{minipage}
	}%
	\subfigure[Adversarial example $ADV$]{
		\label{fig:fl_conv:c}
		\begin{minipage}[t]{0.3\linewidth}
			\centering
			\includegraphics[width=2in]{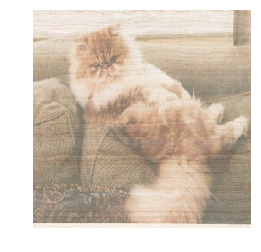}
		\end{minipage}
	}%
	
	\subfigure[Features of $O$]{
		\label{fig:fl_conv:d}
		\begin{minipage}[t]{0.3\linewidth}
			\centering
			\includegraphics[width=2in]{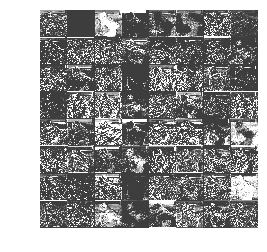}
			
		\end{minipage}
	}%
	\subfigure[Features of $B$]{
		\label{fig:fl_conv:e}
		\begin{minipage}[t]{0.3\linewidth}
			\centering
			\includegraphics[width=2in]{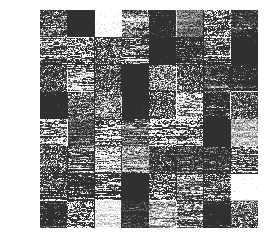}
			
		\end{minipage}
	}%
	\subfigure[Features of $ADV$]{
		\label{fig:fl_conv:f}
		\begin{minipage}[t]{0.3\linewidth}
			\centering
			\includegraphics[width=2in]{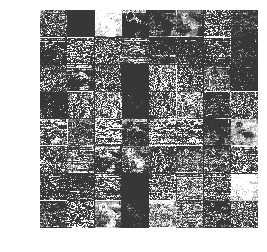}
			
		\end{minipage}
	}%
	
	\centering
	\caption{Features of distorted images extracted by VGG19’s first convolutional layer (after max pooling).After image fusion, the information of $O$ and $B$ is also included in the features of $ADV$, which leads to confusion in feature extraction($\alpha=0.2$). }
	\label{fig:FL_conv_featuremap}
	
\end{figure*}

Although all the above efforts can only solve some problems, chatting is better than nothing.The experimental data show that our defense method improves the defense rate of the model to 80\%.
Our proposed Image Fusion(IF) attack has no effective protection in our known range.From Figure \ref{fig:FL_conv_featuremap},we can find that the features extracted by convolution already contain a large amount information of background image $B$ , and the feature information of the original image has been seriously damaged.Different from noise, it has obvious distribution characteristics different from the original image. The background image is almost perfectly fused with the original image.

\section{Related work}

Previous works mainly study the security and privacy in DL models via white-box mode \cite{szegedy2013intriguing,goodfellow2014explaining,madry2017towards,moosavi2016deepfool}. In the white-box model, the attacker can obtain the adversarial examples quickly and accurately. However, it is difficult for the attacker to know the inner parameters of models in the real world, so  researchers have launched some black-box attacks on  DL models recently such as transfer learning attacks and spatial transformation attacks. 

Transfer learning attacks are first examined by \cite{szegedy2013intriguing}, which study the transferability between different models trained over the same dataset. \cite{Liu2016Delving} propose novel ensemble-based approaches to generate adversarial example . Their approaches enable a large portion of targeted adversarial example to transfer among multiple models for the first time.However, transfer learning attacks have strong limitations, depending on the collection of enough open source models, but for example, there are not enough open source models for pornographic and violent image recognition.

Spatial transformation attacks are very interesting, perturbations generated through spatial transformation could result in large $L_p$ distance measures, but experiments show that such spatially transformed adversarial examples are perceptually realistic and more difficult to defend against with existing defense systems. This potentially provides a new direction in adversarial example generation and the design of corresponding defenses\cite{Xiao2018Spatially}.\cite{Hosseini2017Google} evaluate the robustness of Google Cloud Vision API to input perturbation, they show  that adding an average of 14.25\% impulse noise is enough to deceive the API and when a noise filter is applied on input images, the API generates mostly the same outputs for restored images as for original images.\cite{YuanStealthy} report the first systematic study on the real-world adversarial images and their use in online illicit promotions. \cite{Li2019Adversarial} make the first attempt to conduct an extensive empirical study of black-box attacks against real-world cloud-based image detectors such as violence, politician and pornography detection.
We further extend the spatial transformation attack, and choose the four methods which have the lowest cost to implement. These methods can be implemented without any Deep Learning knowledge and only need a few lines of OpenCV code, but the attack effect is very remarkable. Compared with previous work\cite{Xiao2018Spatially}, these methods are more threatening,we call it \emph{Simple Transformation} (ST) attacks,including \emph{Gaussian Noise}, \emph{Salt-and-Pepper Noise}, \emph{Rotation} and \emph{Monochromatization}.We expand their experiment and use ST to attack models of four cloud-based image classifier services and we propose one novel attack methods, Image Fusion(IF) attack , which achieve a high bypass rate approximately 90\%. Our IF attack can be classified as a spatial transformation attack .Unlike previous work \cite{engstrom2017rotation,Xiao2018Spatially}, we also systematically introduced defense technology and conducted a lot of experiments based on ImageNet datasets.
Experiments show that our defense technology can effectively resist known ST attacks.Through experiments, we prove how to choose different filters in the face of different noises, and how to choose the parameters of different filters.

\section{Conclusion and future work}

In this paper, we mainly focus on studying the security of real-world cloud-based image classifier services. Specifically, We propose one novel attack methods, Image Fusion(IF) attack ,which achieve a high bypass rate ; and we make the first attempt to conduct an extensive empirical study of black-box attacks against real-world cloud-based classifier services. Through evaluations on four popular cloud platforms including Amazon, Google, Microsoft, Clarifai, we demonstrate that ST attack has a success rate of approximately 100\% except Amazon approximately 50\%, IF attack has a success rate over 98\% among different classifier services. 
Finally, we discuss the possible defenses to address these security challenges in cloud-based classifier services.Our defense technology is mainly divided into model training stage and image preprocessing stage.Experiments show that our defense technology can effectively resist known ST attacks, such as  Gaussian Noise, Salt-and-Pepper Noise, Rotation, and Monochromatization.Through experiments, we prove how to choose different filters in the face of different noises, and how to choose the parameters of different filters.

In the future, we aim to explore the space of adversarial examples with less perturbation in black-box and attempt to study targeted attack using IF attack. On the other hand, we will focus on the defense in the cloud environment, so that AI services in the cloud environment away from cybercrime.We hope cloud service providers will not continue to forget this battlefield.



\bibliographystyle{plain}
\bibliography{\jobname}


\appendix

\section*{Appendix}

\begin{figure*}[htbp]
	\centering
	\subfigure[$\alpha=0.2$]{
		\label{fig:fl_cm:a}
		\includegraphics[width=3in]{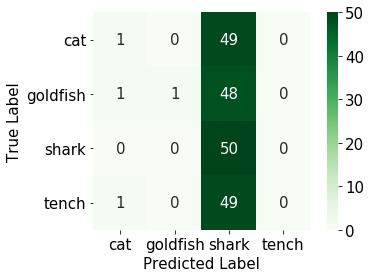}
	}%
	\subfigure[$\alpha=0.4$]{
		\label{fig:fl_cm:b}
		\includegraphics[width=3in]{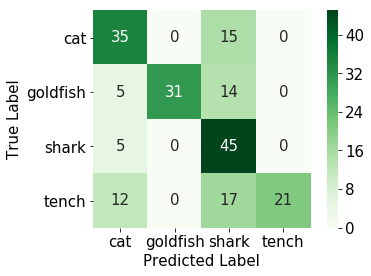}
	}%

	\subfigure[$\alpha=0.6$]{
		\label{fig:fl_cm:c}
		\includegraphics[width=3in]{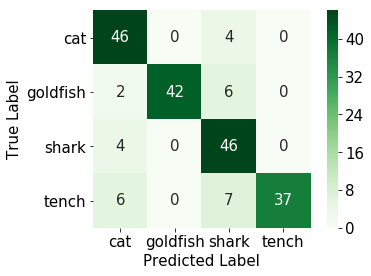}	
	}%
	\subfigure[$\alpha=0.8$]{
		\label{fig:fl_cm:d}
		\includegraphics[width=3in]{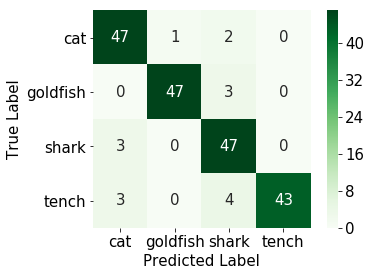}	
	}%

	\subfigure[$\alpha=1.0$]{
		\label{fig:fl_cm:e}
		\includegraphics[width=3in]{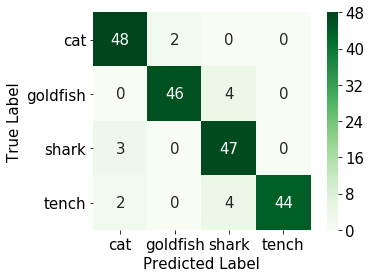}	
	}%
	
	\centering
	\caption{Confusion matrix of \emph{Image Fusion} attack under different parameter $\alpha$.As we can see from the figure, with the increase of fusion parameter $\alpha$, the number of correct classifications increases, and the top-1 accuracy increases. }
\end{figure*}

\begin{figure*}[htbp]
	\subfigure[PSNR]{
		\label{fig:nf_2:c}
		\begin{minipage}[t]{0.5\linewidth}
			\centering
			\includegraphics[width=3in]{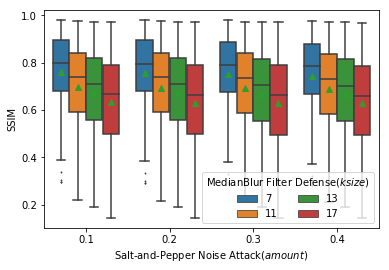}
			
		\end{minipage}
	}%
	\subfigure[SSIM]{
		\label{fig:nf_2:d}
		\begin{minipage}[t]{0.5\linewidth}
			\centering
			\includegraphics[width=3in]{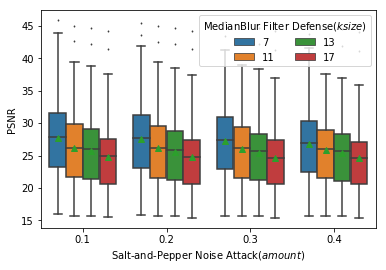}
			
		\end{minipage}
	}%
	\centering
	\caption{We increase $amount$ of \emph{Salt-and-Pepper Noise Attack} from 0.1 to 0.4,  $ksize$ of the \emph{Median Filter} from 3 to 17, the Figure \ref{fig:nf_2:c} records the PSNR  and  the Figure \ref{fig:nf_2:d} records the SSIM  under different parameters.When the parameters $amount$ of \emph{Salt-and-Pepper Noise Attack} are the same, increasing the window size parameters $ksize$ of the \emph{Median Filter} will reduce the image similarity and the image quality.}
	\label{fig:nf_3}
\end{figure*}

\end{document}